%% file: paper.tex
\documentclass[runningheads]{llncs}
\usepackage[T1]{fontenc}
\usepackage{graphicx,verbatim}
\usepackage{hyperref}
\usepackage{tikz}
\usepackage{amsmath}
\usepackage{booktabs}
\usepackage{xcolor}
\usepackage{multirow}
\usepackage{cleveref}
\usepackage[table]{xcolor}
\usepackage{orcidlink}
\newcommand{\ci}[1]{\,{\footnotesize\textcolor{gray}{[#1]}}}

\newcommand{\ciblack}[1]{\,{\textcolor{black}{[#1]}}}

\usepackage{color}

\begin{document}
\title{Foundation Model and Radiomics Distance Scores \\
for Post-Hoc Segmentation Failure Detection}
\titlerunning{FM and Radiomics Distance Scores for Failure Detection}
%

\author{Simon Johannes Joham\inst{1,2}\orcidlink{0009-0009-1817-2904}
\and Gianluca Guglielmo\inst{3}\orcidlink{0009-0001-3446-7528}
\and Mateusz Kozinski\inst{1}\orcidlink{0000-0002-3187-518X} 
\and \\ Martin Urschler\inst{1,2,3}\orcidlink{0000-0001-5792-3971}
}
\authorrunning{S. Joham et al.}
%
\institute{
Institute for Medical Informatics, Statistics and Documentation, \\ Medical University of Graz, Graz, Austria \and
BioTechMed-Graz, Graz, Austria \and
Institute of Visual Computing, Graz University of Technology, Graz, Austria
\email{martin.urschler@medunigraz.at}}
  
\maketitle              
\begin{abstract}
Medical image segmentation models can fail silently when clinical data differs from training data. Deployed models rarely ship with quality control, and are often black boxes. Prior methods that work in this post-hoc failure detection setting still need a separate regression or segmentation model, registration, or access to model internals. We instead propose two failure scores based on the Mahalanobis distance to a labeled reference set. The first uses radiomic shape features of the predicted mask. The second uses frozen foundation model (FM) features, made mask-aware by overlaying the mask on the image, encoding it with a frozen DINOv3, and pooling patch features with more weight on the mask region. We test on multi-site prostate and hippocampus MRI under imaging-protocol and population shift. Ranking cases by predicted failure risk within each site, both scores are competitive with the registration-based (Atlas-RCA) and ensemble-based (pairwise Dice) baselines, without registration that can fail silently or access to individual predictions. Our results show that pooling all test cases into one ranking, as prior work does, rewards ordering sites by difficulty, so a detector can rank sites well, miss failures within them, and still look strong. Within-site evaluation measures true per-case failure. The two evaluation schemes reward different behavior, so we recommend reporting both. The code is
available at \href{https://github.com/imigraz/fm-radiomics-failure-detection}{https://github.com/imigraz/fm-radiomics-failure-detection}. 


\end{abstract}
\section{Introduction}

Reliable segmentation failure detection lets clinicians prioritize the most likely failures for correction~\cite{Zenk2025-FailureDetectSegBenchmark}. Deployed models are often accessible through an API that exposes the predicted segmentation masks, but does not grant users access to model internals. We therefore consider a \emph{post-hoc} failure detection setting: unlike methods that require access to the model's latent features or training pipeline~\cite{Anthony2023-MahalOODNeural,Gonzalez2022-MahalModelOODSameDatasets,Woodland2024-OODFeatures}, we address failure detection for any already-deployed model, based solely on target images and predicted masks. We fit the failure detector once to a labeled reference set where the model works well, in our experiments we use the model training data, and we do not need the reference set at deployment.

Approaches that meet these \emph{post-hoc} constraints fall into three families by the information they use. \textbf{Input-based} methods compare the training and test \emph{image} distributions, e.g.\ via whole-image radiomic features~\cite{Van2017-PyRadiomics} used in the Fr\'echet radiomic distance~(FRD)~\cite{Konz2026-FRD}. \textbf{Output-based} methods read the \emph{mask predictions}, i.e., pairwise Dice score between ensemble predictions~\cite{Roy2019-EnsemblePairwiseMaskDistance}. \textbf{Combined} methods use \emph{both}: reverse classification accuracy~(RCA)~\cite{Valindria2017-AtlasRCA} treats annotated reference images as a prior of plausible anatomy and scores a prediction by how well it transfers to them. Atlas-RCA does this by registration, while ConfIC-RCA~\cite{Cosarinsky2026-ConfIC-RCA} uses an in-context segmentation model with reference retrieval. Each has a setup cost: a specifically trained regression model~\cite{Liu2023-RadQCCardiac,Ribeiro2026-RadQCRegCardiac,Zenk2025-FailureDetectSegBenchmark}, an ensemble of model runs, registration that can fail silently, or a highly capable in-context segmentation model. We instead propose two distance-based scores. Each scores a case by its Mahalanobis distance to a labeled reference set, using only frozen features and no task-specific component: no regression model, no ensemble uncertainty signal, no registration, and no additional segmentation model. 

\begin{figure}[t!]
    \centering
    \includegraphics[width=\textwidth]{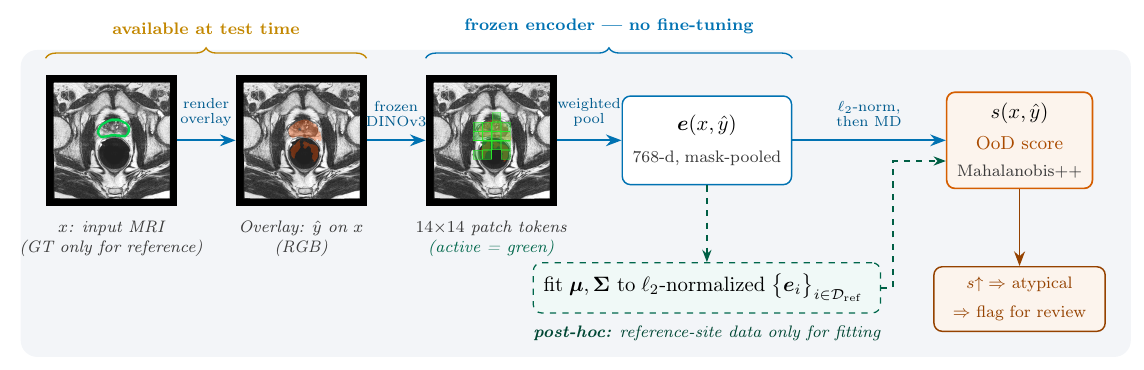}
    \caption{Foundation model score. From the test image $x$ and predicted mask $\hat{y}$, we render an overlay, encode it with a frozen DINOv3, and pool patch features into an embedding $\boldsymbol{e}(x,\hat{y})$, keeping only patches that overlap the mask, and weighting each by its mask coverage. To derive a final per-case segmentation failure signal, the case is scored by the Mahalanobis++ distance $s(x,\hat{y})$ to a reference distribution $(\boldsymbol{\mu},\boldsymbol{\Sigma})$ fit to labeled reference-site cases, without requiring model internals or registration.}
    \label{fig:method_overview}
\end{figure}

Our first proposed score uses \emph{radiomic shape features}~\cite{Van2017-PyRadiomics} of the predicted mask. Prior work~\cite{Liu2023-RadQCCardiac,Ribeiro2026-RadQCRegCardiac,Zenk2025-FailureDetectSegBenchmark} feeds radiomic features to a trained regression model. We hypothesize that the Mahalanobis distance~\cite{lee2018simple} of the radiomic mask shape to the reference shape prior already detects failures well, without a regression model. The second score uses \emph{frozen self-supervised foundation-model~(FM) embeddings}, in particular DINO-family features~\cite{Perez-Garcia2025-RadDino,Simeoni2025-Dinov3,Xu2025-3Dino}, directly as a failure signal. Such embeddings already power anomaly detection~\cite{Schulthess2026-AnomalyDetMedicalImagesDINOv2} or in-context segmentation~\cite{Cuttano2026-Dinov3Seg}, and in quality control~(QC) they serve only for latent-space alignment~\cite{Marciano2025-DiffusionQC} or for reference retrieval~\cite{Cosarinsky2026-ConfIC-RCA}. To our knowledge they have not been used as the failure signal \emph{itself} in a \emph{post-hoc} deployment setting. To derive the FM score, we render the predicted mask onto the image as a colored overlay (see ~\Cref{fig:method_overview}), as in EvanySeg~\cite{Senbi2026-EvanySeg}, but without its fine-tuning of an additional QC model. We pass this overlay through a frozen DINOv3~\cite{Simeoni2025-Dinov3} encoder, and give more weight to patch features that overlap the mask. We fit a reference mean and covariance once on the labeled reference set, then score each test case by its Mahalanobis++~\cite{Mueller2025-Mahalanobispp} distance to that fitted distribution. At deployment only these reference embedding statistics are needed, not the reference images or labels. These statistics can be easily deployed alongside a segmentation model. Beyond DINOv3, we also test RadDINO~\cite{Perez-Garcia2025-RadDino} and 3DINO~\cite{Xu2025-3Dino} features.

We evaluate on multi-site prostate~\cite{Liu2020-SAMLProstate} and hippocampus MRI~\cite{Gonzalez2022-MahalModelOODSameDatasets} under imaging-protocol and population shift, measuring failure detection with the excess area under the risk-coverage curve (E-AURC)~\cite{Geifman2019-EAURC}. Prior work~\cite{Zenk2025-FailureDetectSegBenchmark} pools all site cases into one test set before scoring, which lets a method score well just by ranking easy sites above hard ones, with limited per-case signal. We therefore report both pooled E-AURC, all cases scored together, and mean within-site E-AURC, each site scored on its own and averaged, to isolate per-case detection. Our two proposed scores make this concrete: they tie within-site but split under pooling, which exposes the gap between the two evaluation protocols.

\paragraph{Contribution summary:}
\begin{itemize}
    \item[$\circ$] We propose a model-free radiomic shape feature failure score using Mahalanobis distance. Previous work requires training a regression model.
    \item[$\circ$] We study frozen DINO-family features directly for post-hoc failure detection. When evaluating within-site ranking, the best variant matches the radiomic shape score on both tasks and stays close to best performing registration-based Atlas-RCA and ensemble-based pairwise Dice score. However, the shape and FM scores need no model internals, ensemble uncertainty estimates, in-context segmentation model or registration that can fail silently.
    \item[$\circ$] Our results show that pooled and within-site E-AURC reward different behavior: pooled evaluation, as used in prior work, can reward between-site separability rather than per-case failure detection. We therefore report both.
\end{itemize}

\section{Methods}

\paragraph{Problem setup.}
We address \emph{post-hoc} segmentation failure detection: given a deployed segmentation model, we score each test prediction by its likely failure, without access to the model weights. An adequate failure risk score $s(x,\hat{y})$ should rank cases by segmentation risk so that experts can triage the most severe failures~\cite{Zenk2025-FailureDetectSegBenchmark}. At inference time, only the input image $x$ and its predicted mask $\hat{y}$ are observed, and scoring requires access to the deployed embedding statistics from a reference site, in our case the segmentation model's training data. 
The segmentation model we audit is a five-fold nnU-Net~\cite{Isensee2021-nnU-Net} ensemble trained on a reference dataset. We take the majority-vote mask over the five folds for each test case. Only the ensemble baseline gets access to the individual fold masks.

\paragraph{Distance scoring.}
We ask whether a prediction looks like those the model made on reference cases it handled well, and score how far each case departs from that reference distribution in a feature space $\mathbf{e}$. We compute this with the Mahalanobis distance~(MD)~\cite{Anthony2023-MahalOODNeural,lee2018simple}, a standard Out-of-Distribution~(OoD) detector:
\begin{equation}
    s = (\mathbf{e} - \boldsymbol{\mu})^\top \boldsymbol{\Sigma}^{-1}(\mathbf{e} - \boldsymbol{\mu})\,,
\end{equation}
where $\mathbf{e}(x,m)$ maps an image $x$ and mask $m$ to a feature vector, $\boldsymbol{\mu}$ is the reference mean, and $\boldsymbol{\Sigma}$ the reference covariance. We estimate $\boldsymbol{\Sigma}$ with Ledoit-Wolf shrinkage~\cite{Ledoit2004-LWShrinkage} so the distance stays stable when the feature count is large relative to the reference size. We fit $\boldsymbol{\mu}$ and $\boldsymbol{\Sigma}$ once on all reference (training) case embeddings $\mathbf{e}(x,y)$ using ground-truth masks $y$, so the reference describes correctly segmented anatomy. At test time each query case is scored by $\mathbf{e}(x,\hat{y})$ on its five-fold majority-vote predicted mask $\hat{y}$, so a case is flagged when its features depart from this correctly segmented reference. Both scores plug into this distance and differ only in the feature vector $\mathbf{e}$ and in the normalization choice: the shape score derives $\mathbf{e}$ from the mask alone, the FM score from the image and mask. The FM score $\ell_2$-normalizes the embeddings, a variant known as Mahalanobis++~\cite{Mueller2025-Mahalanobispp}, while the shape score z-scores the raw radiomic features.

\paragraph{Radiomic shape score.}
The first score, which we call Rad-MD-Mask, uses the 3D shape radiomic features~\cite{Van2017-PyRadiomics} of the predicted mask $\hat{y}$ as $\mathbf{e}$. Since we fit the reference on ground-truth labels, the prior describes correct anatomy, so the distance measures how far the predicted geometry departs from plausible shape. We select features on the reference site only, with no query-site data. A first filter drops near-constant features as well as heavy-tailed features, whose excess kurtosis would break the $\chi^2$ behavior the Mahalanobis distance relies on. A second step greedily prunes the most correlated feature until every pair satisfies Pearson $|r| \le 0.90$, which sets the feature count based on the data rather than to a fixed amount. This retains 6 of 14 shape features for prostate and 11 for hippocampus, leaving compact descriptors such as sphericity, flatness, and maximum diameter. The score needs no regression model, unlike prior radiomics-based quality control~\cite{Liu2023-RadQCCardiac,Ribeiro2026-RadQCRegCardiac,Zenk2025-FailureDetectSegBenchmark}.

\paragraph{Foundation models.}
The second score uses frozen foundation model (FM) features. FMs are trained on large image corpora in a self-supervised manner~\cite{li2025survey,Simeoni2025-Dinov3} and encode general features that transfer across domains without fine-tuning. We study whether these features can do failure detection out-of-the-box. This would let practitioners triage segmentations without training a specific QC model, which would require large amounts of labeled training data. We test three FMs: DINOv3 with a ViT-B/16 backbone~\cite{Simeoni2025-Dinov3}, RadDINO with a ViT-B/14 backbone fine-tuned on radiology data~\cite{Perez-Garcia2025-RadDino}, and 3DINO with a 3D ViT backbone trained on 3D CT and MRI~\cite{Xu2025-3Dino}.

\paragraph{Mask-conditioned FM embedding score.}
We obtain $\mathbf{e}$ from an FM encoder, testing input and pooling combinations in~\Cref{tab:data_types}. To build $\boldsymbol{\mu}$ and $\boldsymbol{\Sigma}$, reference-set cases have to be fed through the same encoder. For a 2D backbone, we process the axial slices covering the predicted mask and render each as a grayscale RGB image. For the overlay variants of 2D backbones, we draw the predicted mask on top as a semi-transparent orange overlay (RGB $255,80,0$) at opacity $0.4$. We pass each slice through the frozen encoder and pool its features into one slice embedding in one of two ways: the global variants take the whole-image classification token (CLS) embedding, while the mask-weighted variants pool only the patch features overlapping the mask, weighting each by its mask coverage. For all 2D variants, including the global CLS modes, we embed only axial slices whose predicted-mask area exceeds $10$ pixels, and aggregate the slice embeddings into a case embedding $\boldsymbol{e}(x,\hat{y})$ by weighting each slice by its mask area. For 3DINO, we extract $\mathbf{e}$ directly from 3D MRI. For its overlay variant, we shift the mask-region voxels by $0.5$ in the normalized $[-1,1]$ space (then clip back to $[-1,1]$), since 3DINO is single-channel and cannot take a color overlay. Mask-weighted pooling works the same for 3DINO: we resize the predicted mask to the cubic patch grid and weight each 3D patch token by its mask coverage. Our best-performing DINOv3-MD-OvW, shown in~\Cref{fig:method_overview}, applies the overlay and mask-weighted pooling on a DINOv3 backbone.

\begin{table}[t]
    \footnotesize
    \centering
    \setlength{\tabcolsep}{3pt}
    \caption{Tested embedding variants, spanning two axes: what is used as input (Img, Ov) and how patch features are pooled. For the mask-weighted variants W and OvW, we weight patch features by prediction mask coverage, which gives non-overlapping patches zero weight.}
    \begin{tabular}{lll}
    \toprule
     \textbf{Variant/Mode} & \textbf{Input} & \textbf{Pooling} \\
    \midrule
     Img & Image   & Global (CLS) \\
     Ov  & Image+Overlay & Global (CLS) \\
     W   & Image   & Mask-weighted patches\\
     OvW & Image+Overlay & Mask-weighted patches\\
    \bottomrule
    \end{tabular}
        \label{tab:data_types}
\end{table}

\paragraph{Baselines.}
Ensemble pairwise mask distance (Ens-mpd)~\cite{Roy2019-EnsemblePairwiseMaskDistance} scores each case by the pairwise Dice among the nnU-Net ensemble folds, the best failure-detection method in~\cite{Zenk2025-FailureDetectSegBenchmark}. Atlas-RCA~\cite{Valindria2017-AtlasRCA} registers each test image to all reference images, warps the predicted mask into each reference space, and takes the maximum Dice between the warped prediction and the reference ground truth as the quality score, which still resembles a strong baseline in recent QC literature~\cite{Cosarinsky2026-ConfIC-RCA}. Two further radiomic scores use whole-image features as in FRD~\cite{Konz2026-FRD}: Our Rad-MD-Img scores them with Mahalanobis++, the scorer we also use for the FM embeddings, while Rad-L2~\cite{Konz2026-FRD} uses the L2 distance the authors of FRD suggested as a per-case OoD distance.
\section{Experimental setting}

\input{tables/datasets.tex}

\paragraph{Datasets.}
We adopt the prostate~\cite{Zenk2025-FailureDetectSegBenchmark} and hippocampus~\cite{Gonzalez2022-MahalModelOODSameDatasets} benchmarks from the failure detection literature, both binary segmentation tasks. Contrary to the subset in~\cite{Zenk2025-FailureDetectSegBenchmark}, we retain the original multi-site prostate setup of~\cite{Liu2020-SAMLProstate} and use the preprocessed version of the data from~\cite{Gao2024-Desam}. An overview of the datasets and reference sites is given in~\Cref{tab:datasets}.  We apply isotropic resampling to all images. Each hippocampus is treated as its own case and flipped to the same orientation.

\paragraph{Evaluation protocol.}
We study distance-based failure detection that produces a per-case score, where higher values indicate higher predicted risk. Scores from Atlas-RCA and ensemble-based paired mask distances are negated to follow this convention. In the selective-prediction setting~\cite{Geifman2019-EAURC}, $n$ cases are then ranked by ascending score, and the $k$ lowest-scoring cases are kept at each coverage level $c = k/n$, while the rest are flagged for review. The selective risk at coverage $c$ is the mean risk of the $k$ kept cases, with per-case risk $r = 1 - \text{Dice}$. The area under the risk-coverage curve (AURC)~\cite{Franc2023-FiniteAURC} is the mean selective risk over all coverage levels, which for finite $n$ is
\begin{equation}
    \text{AURC}(s) = \frac{1}{n}\sum_{k=1}^{n} \frac{1}{k}\sum_{i=1}^{k} r_{(i)}  \ ,
\end{equation}
where $r_{(i)}$ is the risk of the case ranked $i$-th by ascending score (lowest predicted risk first). A method that assigns high scores to high-risk cases keeps the retained set clean at every coverage level and yields a low AURC.

We report the gap to the oracle ceiling $\text{AURC}^*$ rather than raw AURC. This keeps the ordering but makes results easier to read, giving the Excess-AURC~\cite{Geifman2019-EAURC},
\begin{equation}
    \text{E-AURC}(s) = \text{AURC}(s) - \text{AURC}^*.
\end{equation}
$\text{AURC}^*$ is the AURC of a perfect failure detection method, so E-AURC reads directly as the performance still to be gained, and removing each site's floor makes per-site values comparable across sites. $\text{E-AURC} = 0$ corresponds to oracle performance (lower is better). We report E-AURC both within-site and pooled. Within-site E-AURC is computed separately per site and averaged with equal weight across sites. Pooling merges all query-site cases into a single ranking, computing E-AURC once. We quantify evaluation uncertainty with a site-stratified percentile bootstrap (2000 iterations, 95\% confidence intervals), method differences are paired for testing. We report E-AURC ($\times10^{-2}$) throughout.

\section{Results}

\input{tables/main_results_tables}

\paragraph{Overall failure detection performance.}
Ranked by within-site mean across both tasks ($\bar{W}$, \Cref{tab:eaurc_mega}), Atlas-RCA leads, our DINOv3-MD-OvW and Ens-mpd follow, and Rad-MD-Mask is close behind. Every other method sits at $\bar{W} \geq 3.20$, so both of our scores approach the level of the two strongest baselines. Within sites, neither score is distinguishable from Atlas-RCA on either task: all four paired differences $\Delta$ (method $-$ reference) stay below $0.8$ E-AURC with CIs spanning zero (Rad-MD-Mask $0.53$\ciblack{-0.76,\,2.22} and $0.72$\ciblack{-0.10,\,1.50}, DINOv3-MD-OvW $0.74$\ciblack{-0.34,\,1.77} and $0.29$\ciblack{-0.45,\,0.98}). Both also match Ens-mpd under protocol shift (CIs spanning zero), while the ensemble keeps a within-site edge under population shift: clear over Rad-MD-Mask ($\Delta{=}1.10$\ciblack{0.56,\,1.64}, $p{=}0.002$) and at the significance boundary over DINOv3-MD-OvW ($\Delta{=}0.67$\ciblack{0.01,\,1.27}, $p{=}0.050$), a small advantage at most. Both of our proposed scores reach this under the strict post-hoc setting, using only reference statistics (mean and covariance matrix) and the deployed model output, whereas Ens-mpd needs access to per-fold masks and Atlas-RCA access to each reference image for registration, which can fail silently. Pooled mean ($\bar{P}$) reorders the leading scores: Ens-mpd and Atlas-RCA lead, Rad-MD-Mask follows, and DINOv3-MD-OvW drops to $\bar{P} = 3.00$.

\paragraph{Pooled vs. within-site.}
Pooled E-AURC can reward two things that within-site ranking does not: separating sites by difficulty, and ranking cases well in the largest site. In prostate, the protocol-shifted sites span Dice 18.0 (BIDMC) to 89.0 (HK, \Cref{tab:datasets}), and Rad-L2 and DINOv3-MD-Img track this site-level risk, scoring better pooled than within-site (5.5 vs. 7.5 and 6.6 vs. 7.3): pooling credits them for splitting easy sites from hard ones, not for ranking cases inside a site. In hippocampus, HarP supplies 270 of 320 cases, so the pooled score is influenced almost entirely by one site and a method can rank the other sites poorly without paying for it. Within-site evaluation removes both effects by ranking cases inside each site, so a method that scores well there is detecting per-case failure, not site structure or size. Atlas-RCA, Rad-MD-Mask, and DINOv3-MD-OvW all stay strong within-site (\Cref{tab:eaurc_mega}), unlike Rad-L2 and DINOv3-MD-Img above. However, pooled evaluation still has its merits, as it reflects a real multi-site triage setting, where flagging the hardest sites is useful, so we report both.

\paragraph{Shape vs. FM features.}
On prostate (protocol shift), Rad-MD-Mask leads pooled ($1.4$ vs.\ $3.8$, $p<0.001$) but the two are indistinguishable within-site ($2.8$ vs.\ $3.0$ with paired $\Delta{=}0.21$\ciblack{-1.93,\,1.52}). On hippocampus (population shift), DINOv3-MD-OvW leads pooled ($2.2$ vs.\ $2.9$, $p=0.025$) and the two again match within-site ($2.2$ vs.\ $2.6$ with paired $\Delta{=}{-}0.43$\ciblack{-1.18,\,0.22}). Only under pooling is the shape score the better signal under protocol shift, and the FM score under population shift. Within-site the two are indistinguishable on both tasks.

\begin{table}[t]
\centering
\small
\setlength{\tabcolsep}{4pt}
\caption{Ablation over backbone (DINOv3, RadDINO, 3DINO) and variants.}
\label{tab:ablation_backbone_mode}
\begin{tabular}{ccccccc}
\toprule
& \multicolumn{2}{c}{\textbf{DINOv3-MD}} & \multicolumn{2}{c}{\textbf{RadDINO-MD}} & \multicolumn{2}{c}{\textbf{3DINO-MD}} \\
\cmidrule(lr){2-3}\cmidrule(lr){4-5}\cmidrule(lr){6-7}
\textbf{Mode} & $\bar{P}$ & $\bar{W}$ & $\bar{P}$ & $\bar{W}$ & $\bar{P}$ & $\bar{W}$ \\
\midrule
Img & 5.25 & 5.10 & 6.15 & 5.30 & 6.25 & 6.45 \\
Ov  & 4.90 & 5.20 & 6.35 & 5.60 & 6.55 & 6.20 \\
W   & 3.95 & 3.20 & \textbf{3.00} & 3.10 & \textbf{2.80} & 3.50 \\
OvW & \textbf{3.00} & \textbf{2.60} & 3.10 & \textbf{3.05} & 3.35 & \textbf{3.40} \\
\bottomrule
\end{tabular}
\end{table}

\paragraph{Image vs. mask-information.}
What the embedding encodes matters more than the choice of FM. Tying the DINOv3 embedding closer to the predicted mask drops the error on prostate within-site, and the other backbones behave similarly (\Cref{tab:ablation_backbone_mode}). A plain whole-image CLS embedding does poorly ($7.3$ E-AURC), as it captures how a site looks, not whether the mask is correct, and adding the overlay alone does not help ($7.7$). Mask-weighted patch pooling drops the error to $4.0$ E-AURC, and adding the overlay on top gives the best score ($3.0$), so the gain comes from steering the embedding toward the mask region. On hippocampus we see the same effect with a smaller spread: its two sites look alike, so even the whole-image embedding works well ($2.9$), yet mask conditioning improves it to $2.2$. Across all three backbones the mask-conditioned modes (W, OvW) beat the whole-image modes (Img, Ov), and no backbone shows a consistent advantage over general DINOv3 within-site across modes (Img, Ov, W, OvW).

\section{Discussion and Conclusion}

Pooling lets a method score well by sorting sites rather than cases, and it weights larger sites more. A detector can then look better when pooled but worse within-site, as Rad-L2 does on prostate. Pooled evaluation is still useful, as it simulates a multi-site annotation triage setting, where sorting sites by difficulty is what matters. Therefore, we report both evaluation schemes, and we recommend the same for future failure-detection work. The two feature families tie within sites on both tasks. They behave differently only under pooling, where shape leads on prostate and FM on hippocampus. Since pooling rewards site ordering, we do not treat this split as per-case performance evidence. Future work should test whether unifying the two families improves per-case failure detection.

\paragraph{Limitations.} The study covers binary segmentation of two anatomies on 3D MRI, with a single reference site and a single segmentation model per task. We did not test sensitivity to the choice of reference set, which in principle only needs cases the model segments well, not the training data specifically. Confirming that a held-out or deployment-site reference gives comparable detection is important future work. Furthermore, more anatomically complex or multi-class targets, such as lesion segmentation, may shift the balance toward appearance cues, since lesions lack the consistent shape prior that organs provide.

\paragraph{Conclusion.}
We studied two distance-based scores for post-hoc segmentation failure detection, one using shape, the other FM features. Both need no model internals, ensemble uncertainty estimates, or registration. Across two domain-shifted multi-site MRI tasks, the predicted mask is the strongest cue for failure detection. The model-free radiomic shape score detects failures competitively, and an FM matches it within sites, without fine-tuning, once its patch embeddings are conditioned on the predicted mask. The two scores behave differently only under pooled evaluation. However, our main finding is that pooled and within-site evaluation reward different behavior (multi-site triage vs. true per-case detection), so ranking detectors fairly needs both.

    

\begin{credits}
\subsubsection{\ackname} This research was funded in whole or in part by the Austrian Science Fund (FWF) 10.55776/PAT1748423. Some computational results have been achieved using the Austrian Scientific Computing (ASC) infrastructure.

\subsubsection{\discintname}
The authors have no competing interests to declare that are
relevant to the content of this article.
\end{credits}

%
%
%
\bibliographystyle{splncs04}
\bibliography{refs.bib}
\end{document}

%% file: tables/datasets.tex
\begin{table}[t!]
\setlength{\tabcolsep}{3pt}
\caption{Dataset summary. Each hippocampus is treated as its own case and flipped to same orientation. nnU-Net ensembles are trained on the reference site (ref.) only, all other sites are held out for evaluation. Ref. Dice column shows mean test performance.}  
\label{tab:datasets}
\footnotesize
\centering
\begin{tabular}{lllll}
\toprule
\textbf{Setting}& \textbf{Site} & $\mathbf{N}$ & \textbf{Description} & \textbf{Ref. Dice} \\
\midrule
\multirow{6}{*}{\textbf{Prostate}~\cite{Liu2020-SAMLProstate}}
 & RUNMC~\cite{Bloch2015-NCI-ISBIProstDataset} & $30$ & Surface coil, 3T (ref.)         & $92.8 \pm 2.0$ \\
 & BMC~\cite{Bloch2015-NCI-ISBIProstDataset}   & $30$ & Endorectal coil, 1.5T           & $63.1 \pm 22.4$ \\
 & I2CVB~\cite{Lemaitre2015-I2CVBProstDataset} & $19$ & No coil, 3T                     & $68.2 \pm 16.4$ \\
 & UCL~\cite{Litjens2014-PROMISE12ProstDataset}   & $13$ & No coil, 1.5--3T                & $86.0 \pm 3.80$ \\
 & BIDMC~\cite{Litjens2014-PROMISE12ProstDataset} & $12$ & Endorectal coil, 3T & $18.0 \pm 19.0$ \\
 & HK ~\cite{Litjens2014-PROMISE12ProstDataset}   & $12$ & Endorectal coil, 1.5T  & $89.0 \pm 2.60$ \\
\midrule
\multirow{3}{*}{\textbf{Hippocampus}~\cite{Gonzalez2022-MahalModelOODSameDatasets}}
 & DecathHip~\cite{Antonelli2022-Decathlon} & $260$ & Healthy (ref.)         & $89.6 \pm 3.3$\\
 & Dryad~\cite{Kulaga-Yoskovitz2015-Dryad}     &  $50$ & Schizophr. \& healthy    & $78.9 \pm 2.6$ \\
 & HarP~\cite{Boccardi2015-HarP}     & $270$ & Alzheimer's \& elderly     & $54.6 \pm 15.4$\\
\bottomrule
\end{tabular}
\end{table}

%% file: tables/main_results_tables.tex
\begin{table}[t]
\caption{Pooled and within-site mean E-AURC ($\times10^{-2}$ $\downarrow$; 95\% CI), for prostate ($n{=}86$, 5 sites) and hippocampus ($n{=}320$, 2 sites). $\bar{P}$ and $\bar{W}$ are the pooled and within-site arithmetic means taken over the two settings. \textbf{Bold}/\underline{underline}: best/second-best per column. Ens-Mpd needs access to ensemble predictions. Atlas-RCA needs registration on reference set. The best proposed scores are highlighted with gray background.}
\setlength{\tabcolsep}{1pt}
\label{tab:eaurc_mega}
\centering\footnotesize
\begin{tabular}{lccccccc}
\toprule
 & & \multicolumn{2}{c}{\textbf{Prostate}} & \multicolumn{2}{c}{\textbf{Hippocampus}} & \multirow{2}{*}{$\bar{P}$} & \multirow{2}{*}{$\bar{W}$} \\
\cmidrule(lr){3-4}\cmidrule(lr){5-6}
\textbf{Method} & \textbf{Mode} & Pooled & Within-site & Pooled & Within-site & & \\
\midrule
Ens-Mpd~\cite{Roy2019-EnsemblePairwiseMaskDistance} & Mask & \underline{1.6}\ci{0.9,2.4} & 3.7\ci{1.4,5.7} & \textbf{1.5}\ci{1.1,1.9} & \textbf{1.5}\ci{1.1,2.0} & \textbf{1.55} & \underline{2.60} \\
Atlas-RCA~\cite{Valindria2017-AtlasRCA} & Img,Mask & 1.7\ci{1.2,2.2} & \textbf{2.2}\ci{1.2,3.6} & \underline{1.8}\ci{1.2,2.7} & \underline{1.9}\ci{1.3,2.5} & \underline{1.75} & \textbf{2.05} \\
Rad-L2~\cite{Konz2026-FRD} & Img & 5.5\ci{3.9,7.2} & 7.5\ci{4.5,9.8} & 8.6\ci{7.3,9.9} & 6.2\ci{5.3,6.9} & 7.05 & 6.85 \\
\midrule
Rad-MD & Img & 6.6\ci{4.7,8.8} & 5.6\ci{3.9,7.1} & 3.7\ci{3.1,4.3} & 4.0\ci{3.4,4.6} & 5.15 & 4.80 \\
\rowcolor{gray!20}
Rad-MD & Mask & \textbf{1.4}\ci{1.0,1.7} & \underline{2.8}\ci{1.3,4.3} & 2.9\ci{2.3,3.6} & 2.6\ci{2.1,3.2} & 2.15 & 2.70 \\
DINOv3-MD & Img & 6.6\ci{4.6,8.6} & 7.3\ci{3.9,9.8} & 3.9\ci{3.3,4.6} & 2.9\ci{2.3,3.4} & 5.25 & 5.10 \\
DINOv3-MD & Ov & 6.3\ci{4.5,8.0} & 7.7\ci{4.2,10.2} & 3.5\ci{3.0,4.1} & 2.7\ci{2.2,3.2} & 4.90 & 5.20 \\
DINOv3-MD & W & 5.0\ci{3.4,6.7} & 4.0\ci{2.1,5.8} & 2.9\ci{2.4,3.4} & 2.4\ci{1.9,2.8} & 3.95 & 3.20 \\
\rowcolor{gray!20}
DINOv3-MD & OvW & 3.8\ci{2.8,4.9} & 3.0\ci{1.5,4.2} & 2.2\ci{1.8,2.5} & 2.2\ci{1.7,2.6} & 3.00 & \underline{2.60} \\
\bottomrule
\end{tabular}
\end{table}